\def\year{2018}\relax
\documentclass[letterpaper]{article} 
\usepackage{aaai18}  
\usepackage{times}  
\usepackage{helvet}  
\usepackage{courier}  
\usepackage{url}  
\usepackage{graphicx}  

\usepackage{booktabs}       
\usepackage{amsfonts}       
\usepackage{nicefrac}       
\usepackage{microtype}      
\usepackage{times}
\usepackage{latexsym}
\usepackage{tikz}
\usetikzlibrary{arrows}
\usepackage{amsmath} 
\usepackage{pifont}

\usetikzlibrary{matrix}
\usetikzlibrary{positioning}
\usepackage{array}
\usepackage{multirow}
\usepackage{enumitem}
\usepackage{booktabs}
\usepackage{subcaption}
\usepackage{placeins}

\newcommand{\citet}[1]{\citeauthor{#1}~\shortcite{#1}}
\newcommand{\citep}{\cite}

\begin{document}
%
\title{Sentence Ordering and Coherence Modeling using Recurrent Neural Networks}

\author{Lajanugen Logeswaran$^1$, Honglak Lee$^1$, Dragomir Radev$^2$ \\
$^1$Department of Computer Science \& Engineering, University of Michigan \\
$^2$Department of Computer Science, Yale University \\
\textit{llajan@umich.edu, honglak@eecs.umich.edu, dragomir.radev@yale.edu}
}

\maketitle
\begin{abstract}

Modeling the structure of coherent texts is a key NLP problem.
The task of coherently organizing a given set of sentences has been commonly used to build and evaluate models that understand such structure.
We propose an end-to-end unsupervised deep learning approach based on the set-to-sequence framework to address this problem.
Our model strongly outperforms prior methods in the order discrimination task and a novel task of ordering abstracts from scientific articles.
Furthermore, our work shows that useful text representations can be obtained by learning to order sentences.
Visualizing the learned sentence representations shows that the model captures high-level logical structure in paragraphs.
Our representations perform comparably to state-of-the-art pre-training methods on sentence similarity and paraphrase detection tasks.
\end{abstract}

\section{Introduction}
\label{intro}
Modeling the structure of coherent texts is an important problem in NLP. 
A well-written text has a particular high-level logical and topical structure. 
The actual word and sentence choices and their transitions come together to convey the purpose of the text.
Our primary goal is to build models that can learn such structure by arranging a given set of sentences to make coherent text. 


Multi-document Summarization (MDS) and retrieval-based Question Answering (QA) involve extracting information from multiple documents and organizing it into a coherent summary.
Since the relative ordering of sentences from different sources can be unclear, being able to automatically evaluate a particular order is essential.
\citet{barzilay2002inferring} discuss the importance of an ordering component in MDS and show that finding acceptable orderings can enhance user comprehension.

More importantly, by learning to order sentences we can model text coherence.
It is difficult to explicitly characterize the properties of text that make it coherent.
Ordering models attempt to understand these properties by learning high-level structure that causes sentences to appear in a specific order in human-authored texts. 
Automatic methods for evaluating human/machine generated text have great importance, with 
applications in essay scoring \citep{miltsakaki2004evaluation,burstein2010using} and text generation \citep{park2015expressing,kiddon2016globally}.
Coherence models aid the better design of these systems.

Exploiting unlabelled corpora to learn semantic representations of data has become an active area of investigation. 
Self-supervised learning is a typical approach that uses information naturally available as part of the data as supervisory signals \citep{wang2015unsupervised,doersch2015unsupervised}.
\citet{noroozi2016unsupervised} attempt to learn visual representations by solving image jigsaw puzzles.
Sentence ordering can be considered as a jigsaw puzzle in the language domain and an interesting question is whether we can learn useful textual representations by performing this task.



Our approach to coherence modeling is driven by recent success in capturing semantics using distributed representations and modeling sequences using Recurrent Neural Nets (RNN).
RNNs are now the dominant approach to sequence learning and mapping problems.
The Sequence-to-sequence (Seq2seq) framework \citep{sutskever2014sequence} and its variants have fueled RNN based approaches to a range of problems such as language modeling, text generation, MT, QA, and many others.

In this work we propose an RNN-based approach to the sentence ordering problem which exploits the set-to-sequence framework of \citet{vinyals2015order}.
A word-level RNN encoder produces sentence embeddings, and a sentence-level set encoder RNN iteratively attends to these embeddings and constructs a context representation.
Initialized with this representation, a sentence-level pointer network selects the sentences sequentially.

The most widely studied task relevant to sentence ordering and coherence modeling is the order discrimination task.
Given a document and a permuted version of it, the task involves identifying the more coherent ordering. 
Our proposed model achieves state-of-the-art performance on two benchmark datasets for this task, outperforming several classical approaches and recent data-driven approaches.

Addressing the more challenging task of ordering a given collection of sentences, we consider the novel and interesting task of ordering sentences from abstracts of scientific articles.
Our model strongly outperforms previous work on this task.
We visualize the learned sentence representations and show that our model captures high-level discourse structure.
We provide visualizations that help understand what information in the sentences the model uses to identify the next sentence. 

Finally, we demonstrate that our ordering model learns coherence properties and text representations that are useful for several downstream tasks including summarization, sentence similarity and paraphrase detection.
In summary, our key contributions are as follows:
\begin{itemize}[leftmargin=*]
	\item We propose an end-to-end trainable model based on the set-to-sequence framework to address the problem of coherently ordering a collection of sentences.
    \item We consider the novel task of understanding structure in abstract paragraphs and demonstrate state-of-the-art results in order discrimination and sentence ordering tasks.
    \item We show that our model learns sentence representations that perform comparably to recent unsupervised pre-training methods on downstream tasks.
\end{itemize}




\section{Related Work}
\label{related}
\textbf{Coherence modeling \& sentence ordering.} 
Coherence modeling and sentence ordering have been approached by closely related techniques.
Many approaches propose a measure of coherence and formulate the ordering problem as finding an order with maximal coherence.
Recurring themes from prior work include linguistic features, centering theory, local and global coherence.  
Local coherence has been modeled by considering properties of local windows of sentences such as sentence similarity and transition structure.
\citet{lapata2003probabilistic} represent sentences by vectors of linguistic features and learn the transition probabilities between features of adjacent sentences.
The Entity-Grid model \cite{barzilay2008modeling} captures local coherence by modeling patterns of entity distributions. 
Sentences are represented by the syntactic roles of entities appearing in the document.
Features extracted from the entity grid are used to train a ranking SVM.
These two methods are motivated from centering theory \citep{grosz1995centering}, which states that nouns and entities in coherent discourses exhibit certain patterns.


Global models of coherence typically use HMMs to model document structure.
The content model \cite{barzilay2004catching} represents topics in a particular domain as states in an HMM.
State transitions capture possible presentation orderings within the domain.
Words of a sentence are modeled using a topic-specific language model.
The content model inspired several subsequent work to combine the strengths of local and global models.
\citet{elsner2007unified} combine the entity grid and the content model using a non-parametric HMM.
\citet{soricut2006discourse} use several models as feature functions and define a log-linear model to assign probability to a text.
\citet{louis2012coherence} model the intentional structure in documents using syntax features. 

Unlike previous approaches, we do not use any handcrafted features and adopt an embedding-based approach.
Local coherence is taken into account by a next-sentence prediction component in our model, and global dependencies are naturally captured by an RNN. 
We demonstrate that our model can capture both logical and topical structure by several evaluation benchmarks. 

\textbf{Data-driven approaches.}
Neural approaches have gained attention recently.
\citet{li2014model} model sentences as embeddings derived from recurrent neural nets and train a feed-forward neural network that takes an input window of sentence embeddings  and outputs a probability which represents the coherence of the sentence window.
Coherence evaluation is performed by sliding the window over the text and aggregating the score.
\citet{li2016neural} study the same model in a larger scale task and also use a sequence-to-sequence approach in which the model is trained to generate the next sentence given the current sentence and vice versa.
\citet{nguyen2017neural} learn to model coherence using a convolutional network that operates on the Entity-Grid representation of an input document.
These models are limited by their local nature; our experiments show that considering larger contexts is beneficial.

\textbf{Hierarchical RNNs for document modeling.}
Word-level and sentence-level RNNs have been used in a hierarchical fashion for modeling documents in prior work. 
\citet{li2015hierarchical} proposed a hierarchical autoencoder for generation and summarization applications.
More relevant to our work is a similar model considered by \citet{lin2015hierarchical}. 
A sentence-level RNN predicts the bag of words in the next sentence given the previous sentences 
and a word-level RNN predicts the word sequence conditioned on the sentence RNN hidden state.
Our model has a hierarchical structure similar to these models, but takes a discriminative approach. 

\textbf{Combinatorial optimization with RNNs.}
\citet{vinyals2015order} equip sequence-to-sequence models with the ability to handle input and output sets, and discuss experiments on sorting, language modeling and parsing.
This is called the \textit{read}, \textit{process} and \textit{write} (or set-to-sequence) model.
The \textit{read} block maps input tokens to a fixed length vector representation.
The \textit{process} block is an RNN encoder which, at each time-step, attends to the input token embeddings and computes an attention readout, appending it to the current hidden state.
The \textit{write} block is an RNN which produces the target sequence conditioned on the representation produced by the process block.
Their goal is to show that input and output orderings can matter in these tasks, which is demonstrated using small scale experiments. 
Our work exploits this framework to address the challenging problem of modeling logical and hierarchical structure in text.
\citet{vinyals2015pointer} proposed pointer-networks for combinatorial optimization problems where the output dictionary size depends on the number of input elements.
We use a pointer-network as the decoder to sequentially pick the next sentence.

\vspace{-0.7em}
\section{Approach}
\label{sec:approach}
Our proposed model is inspired by the way a human would solve this task.
First, the model reads the sentences to capture their meaning and the general context of the paragraph.
Given this knowledge, the model tries to pick the sentences one by one sequentially till exhaustion.

Our model is based on the \textit{read}, \textit{process} and \textit{write} framework of \citet{vinyals2015order} briefly discussed in the previous section.
We use the encoder-decoder terminology that is more common in the following discussion.


\begin{figure*}[t]
\center
\includegraphics[width=1.0\textwidth]{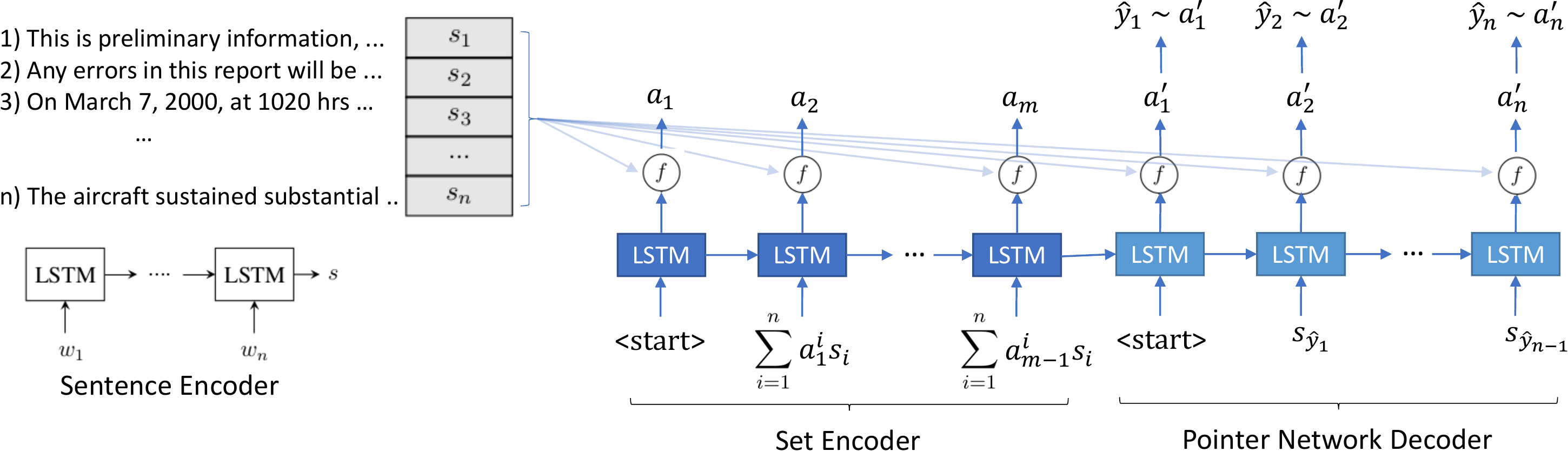}
\caption{\textbf{Model Overview}: 
The input set of sentences are represented as vectors using a sentence encoder.
At each time step of the model, attention weights are computed for the sentence embeddings based on the current hidden state. 
The encoder uses the attention probabilities to compute the input for the next time-step and the decoder uses them for prediction.
}
\label{arch}
\end{figure*}

The model is comprised of a sentence encoder RNN, an encoder RNN and a decoder RNN (Fig. \ref{arch}).
The sentence encoder takes as input the words of a sentence $s$ sequentially and computes an embedding representation of the sentence. 
Henceforth, we use $s$ to refer to a sentence or its embedding interchangeably.
The embeddings $\{s_1,s_2,...,s_n\}$ of a given set of $n$ sentences constitute the sentence memory, available to be accessed by subsequent components.

The encoder LSTM is identical to the originally proposed process block, defined by Eqs \ref{enc1}-\ref{enc4}. 
At each time step the input to the LSTM is computed by taking a weighted sum over the memory elements, the weights being attention probabilities obtained using the previous hidden state as query (Eqs. \ref{enc1}, \ref{enc2}).
This is iterated for a fixed number of times called the read cycles.
Intuitively, the model identifies a soft input order to read the sentences.
As described in \citet{vinyals2015order} the encoder has the desirable property of being invariant to the order in which the sentence embeddings reside in the memory.
\begin{align}
	e^{t,i}_{\text{enc}} &= f(s_i, h^{t}_{\text{enc}}) ; i\in\{1,...,n\} \label{enc1} \\
	a^t_\text{enc} &= \text{Softmax}(e^t_{\text{enc}}) \label{enc2} \\
	s^t_{\text{att}} &= \sum_{i=1}^n a^{t,i}_\text{enc} s_i  \\
	h^{t+1}_{\text{enc}}, c^{t+1}_{\text{enc}} &= \text{LSTM}(h^{t}_{\text{enc}}, c^{t}_{\text{enc}}, s^{t}_{\text{att}})  \label{enc4}  
\end{align}
The decoder is a pointer network that takes a similar form with a few differences (Eqs. \ref{dec1}-\ref{dec4}). 
The LSTM takes the embedding of the previous sentence as input instead of the attention readout.
At training time the correct order of sentences $(s_{o_1}, s_{o_2},...,s_{o_n}) = (x^1, x^2,...,x^n)$ is known ($o$ represents the correct order) and $x^{t-1}$ is used as the input.
At test time the predicted assignment $\hat{x}^{t-1}$ is used instead.
The attention computation is identical to that of the encoder, but now $a^{t,i}_{\text{dec}}$ is interpreted as the probability for $s_i$ being the correct sentence choice at position $t$, conditioned on the previous sentence assignments $p(S_{t} = s_i|S_1,...,S_{t-1})$.
The initial state of the decoder LSTM is initialized with the final hidden state of the encoder. 
$x^0$ is a vector of zeros. 

\abovedisplayskip=-10pt
\begin{align}
	h^{t}_{\text{dec}}, c^{t}_{\text{dec}} &= \text{LSTM}(h^{t-1}_{\text{dec}}, c^{t-1}_{\text{dec}}, x^{t-1}) \label{dec1} \\
	e^{t,i}_{\text{dec}} &= f(s_i, h^t_{\text{dec}}) ; i\in\{1,...,n\} \label{dec3} \\
	a^t_\text{dec} &= \text{Softmax}(e^t_{\text{dec}}) \label{dec4}
\end{align}
\abovedisplayskip 7pt plus2pt minus5pt
\textbf{Scoring Function}. 
We consider two choices for the scoring function $f$ in Eqs. \ref{enc1}, \ref{dec3}. 
The first (Eq. \ref{eq:mlp}) is a single hidden layer feed-forward net that takes $s,h$ as inputs ($W, b, W', b'$ are learnable parameters). 
The structure of $f$ is similar to the window network of \citet{li2014model}.
While they used a local window of sentences to capture context, this scoring function exploits the entire history of sentences encoded in the RNN hidden state to score candidates for the next sentence. 
\begin{equation}
  	f(s,h) = W' \text{tanh}(W[s\thinspace;\thinspace h] + b) + b'
  	\label{eq:mlp}
\end{equation} 

We also consider a bilinear scoring function (Eq. \ref{eq:bilin}). 
Compared to the previous scoring function, this takes a generative approach to regress the next sentence given the current hidden state $(Wh+b)$, enforcing that it be most similar to the correct next sentence. 
We observed that this scoring function led to better sentence representations (Sec. \ref{senrep}).
\begin{equation}
  	f(s,h) = s^T (W h + b)
  	\label{eq:bilin}
\end{equation}

\textbf{Contrastive Sentences}.
In its vanilla form, we found that the set-to-sequence model tends to rely on certain word clues to perform the ordering task. 
To encourage holistic sentence understanding, we add a random set of sentences to the sentence memory when the decoder makes classification decisions.
This makes the problem more challenging for the decoder since now it has to distinguish between sentences that are relevant and irrelevant to the current context in identifying the correct sentence.

\begin{table*}[!t]
\centering
	\caption{Mean Accuracy comparison on the Accidents and Earthquakes data for the order discrimination task. The reference models are Entity-Grid \citep{barzilay2008modeling}, HMM \citep{louis2012coherence},  Graph \citep{guinaudeau2013graph},  Window network \citep{li2014model} and sequence-to-sequence \citep{li2016neural}, respectively. 
	}
	\begin{tabular}{lccccccc} \toprule
	& Entity-Grid & HMM & Graph &  \multicolumn{2}{c}{Window} & Seq2seq & Ours \\
	& & & & (Recurrent) & (Recursive) & & \\
	\hline \\[-0.8em]
	Accidents   & 0.904 & 0.842 & 0.846 & 0.840 & 0.864 & 0.930 & \textbf{0.944} \\
	Earthquakes & 0.872 & 0.957 & 0.635 & 0.951 & 0.976 & 0.992 & \textbf{0.997} \\
	\bottomrule
	\end{tabular}
	\label{table:discrimination} 
\end{table*}

\textbf{Coherence modeling}.
We define the coherence score of an arbitrary partial/complete assignment $(s_{p_1},...,s_{p_k})$ to the first $k$ sentence positions as
\begin{equation}
	\textstyle{\sum}_{i=1}^k \text{log }
	p(S_{i}=s_{p_{i}}|S_{1,...,i-1}=s_{p_{1},...,p_{i-1}})
\end{equation}
where $S_1,..,S_k$ are random variables representing the sentence assignment to positions $1$ through $k$.  
The conditional probabilities are derived from the network.
This is our measure of comparing the coherence of different renderings of a document.
It is also used as a heuristic during decoding.

\textbf{Training Objective}.
The model is trained using the maximum likelihood objective 
\begin{equation}
	\text{max} \textstyle{\sum}_{x\in D} \textstyle{\sum}_{t=1}^{|x|} \text{log } p(x^t|x^1,...,x^{t-1})
	\label{eq:loss}                                          
\end{equation}
where $D$ denotes the training set and each training instance is given by an ordered document of sentences $x = (x^1,...,x^{|x|})$.



\section{Experimental Results}

We first consider the order discrmination task that has been widely used in the literature for evaluating coherence models.
We then consider the more challenging ordering problem where a coherent order of a given collection of sentences needs to be determined.
We then demonstrate that our ordering model learns coherence properties useful for summarization.
Finally, we show that our model learns sentence representations that are useful for downstream applications.

For all tasks discussed in this section we train the model with the maximum likelihood objective on the training data relevant to the task.
We used the single hidden layer MLP scoring function for the order discrimination and sentence ordering tasks.
Models are trained end-to-end. 
We use pre-trained 300 dimensional GloVe word embeddings~\citep{pennington2014glove} to initialize word vectors.
All LSTMs use a hidden layer size of 1000 and the MLP in Eq.~\ref{eq:mlp} has a hidden layer size of 500.
The number of read cycles in the encoder is set to 10. 
The same model architecture is used across all experiments.
We used the Adam optimizer \citep{kingma2014adam} with batch size 10 and learning rate 5e-4 for learning. 
The model is regularized using early stopping.
Hyperparameters were chosen using the validation set.

\subsection{Order Discrimination}
The ordering problem is traditionally formulated as a binary classification task:
Given a reference paragraph and its permuted version, identify the more coherent one \citep{barzilay2008modeling}.


The datasets widely used for this task in previous work are the Accidents and Earthquakes news reports.
In each of these datasets the training and test sets include 100 articles and approximately 20 permutations of each article.

\begin{table*}[ht!]
\centering
\label{Sentence Reordering}
\caption{Comparison against prior methods on the abstracts data.}
\begin{tabular}{lcccccc} \toprule
\multicolumn{1}{c}{} & \multicolumn{2}{c}{NIPS Abstracts} & \multicolumn{2}{c}{AAN Abstracts} & \multicolumn{2}{c}{NSF Abstracts} \\
\cmidrule(l{2pt}r{2pt}){2-3} \cmidrule(l{2pt}r{2pt}){4-5} \cmidrule(l{2pt}r{2pt}){6-7}
\multicolumn{1}{c}{} & \multicolumn{1}{c}{Accuracy} & \multicolumn{1}{c}{$\tau$} & \multicolumn{1}{c}{Accuracy} & \multicolumn{1}{c}{$\tau$}  & \multicolumn{1}{c}{Accuracy} & \multicolumn{1}{c}{$\tau$} \\
\cmidrule{1-7}
Random & 15.59 & 0 & 19.36 & 0 & 9.46 & 0 \\
\cmidrule{1-7}
Entity Grid \citep{barzilay2008modeling} & 20.10 & 0.09 & 21.82 & 0.10 & - & - \\
Seq2seq (Uni) \citep{li2016neural} & 27.18 & 0.27 & 36.62 & 0.40 & 13.68 & 0.10 \\ 
Window network \citep{li2014model}			& 41.76 & 0.59 & 50.87 & 0.65 & 18.67 & 0.28 \\
RNN Decoder & 48.22 & 0.67 & 52.06 & 0.66 & 25.79 & 0.48 \\
Proposed model & \textbf{51.55} & \textbf{0.72}  & \textbf{58.06}  & \textbf{0.73}  & \textbf{28.33}  & \textbf{0.51}  \\
\bottomrule
\end{tabular}

\label{nips}
\end{table*}

In Table \ref{table:discrimination} we compare our results with traditional approaches and recent data-driven approaches. 
The entity grid model provides a strong baseline on the Accidents dataset, only outperformed by our model and \citet{li2016neural}.
On the Earthquakes data the window approach of \citet{li2016neural} performs strongly.
Our approach outperforms prior models on both datasets, achieving near perfect performance on the Earthquakes dataset.

While these datasets have been widely used, they are quite formulaic in nature and are no longer challenging. 
We hence turn to the more challenging task of ordering a given collection of sentences to make a coherent document. 

\subsection{Sentence Ordering}
In this task we directly address the ordering problem. 
We do not assume the availability of a set of candidate orderings to choose from and instead find a good ordering from all possible permutations of the sentences.

The difficulty of the ordering problem depends on the nature of the text, as well as the length of paragraphs considered. 
Evaluation on text from arbitrary text sources makes it difficult to interpret the results, since it may not be clear whether to attribute the observed performance to a deficient model or ambiguity in next sentence choices due to many plausible orderings.

Text summaries are a suitable source of data for this task.
They often exhibit a clear flow of ideas and have little redundancy.
We specifically look at abstracts of conference papers and research proposals. 
This data has several favorable properties.
Abstracts usually have a particular high-level format - They begin with a brief introduction, a description of the problem and proposed approach and conclude with performance remarks.
This would allow us to identify if the model can capture high-level logical structure.
Second, abstracts have an average length of about 10, making the ordering task more accessible.
This also gives us a significant amount of data to train and test our models.


We use the following sources of abstracts for this task. 
\begin{itemize}
	\item \textit{NIPS Abstracts}. 
We consider abstracts from NIPS papers in the past 10 years.
We parsed 3280 abstracts from paper pdfs and obtained 3259 abstracts after omitting erroneous extracts.
The dataset was split into years 2005-2013 for training and 2014, 2015 respectively for validation, testing. 

\item \textit{ACL Abstracts}. A second source of abstracts are papers from the ACL Anthology Network (AAN) corpus~\citep{Radev&al.09}. 
We extracted 12,157 abstracts from the text parses using simple keyword matching for the strings `Abstract' and `Introduction'.
We use all extracts of papers published up to year 2010 for training, year 2011 for validation and years 2012-2013 for testing. 

\item \textit{NSF Abstracts}. 
We also used the NSF Research Award Abstracts dataset~\citep{Lichman:2013}.
It comprises abstracts from a diverse set of scientific areas in contrast to the previous two sources of data and the abstracts are also lengthier, making this dataset more challenging.
Years 1990-1999 were used for training, 2000 for validation and 2001-2003 for testing.
We capped the parses of the abstracts to a maximum length of 40 sentences.
Unsuccessful parses and parses of excessive length were discarded.
\end{itemize}
Further details about the data are provided in the supplement.

The following metrics are used to evaluate performance.
\textit{Accuracy} measures how often the absolute position of a sentence was correctly predicted.
\textit{Kendall's tau ($\tau$)} is computed as $1 - 2\cdot N / {n \choose 2}$, where $N$ is the number of pairs in the predicted sequence with incorrect relative order and $n$ is the sequence length.
\citet{lapata2006automatic} discusses that this metric reliably correlates with human judgements. 

The following baselines are used for comparison:
\begin{itemize}
	\item \textbf{Entity Grid}. 
Our first baseline is the Entity Grid model of \citet{barzilay2008modeling}.
We use the Stanford parser~\citep{klein2003accurate} and Brown Coherence Toolkit\footnote{\url{bitbucket.org/melsner/browncoherence}} to derive Entity grid representations. 
A ranking SVM is trained to score correct orderings higher than incorrect orderings as in the original work.
We used 20 permutations per document as training data.
Since the entity grid only provides a means of feature extraction we evaluate the model in the ordering setting as follows.
We choose 1000 random permutations for each document, one of them being the correct order, and pick the order with maximum coherence.
We experimented with transitions of length at most 3 in the entity-grid.

\item \textbf{Seq2seq}. The second baseline we consider is a sequence-to-sequence model which is trained to predict the next sentence given the current sentence.
\citet{li2016neural} consider similar methods and our model is the same as their uni-directional model.
These methods were shown to yield sentence embeddings that have competitive performance in several semantic tasks in \citet{kiros2015skip}.

\item \textbf{Window Network}. We consider the window approach of \citet{li2014model} and \citet{li2016neural} which demonstrated strong performance in the order discrimination task as our third baseline. 
We adopt the same coherence score interpretation considered by the authors. 
In both the above models we consider a special embedding vector which is padded at the beginning of a paragraph and learned during training.
This vector allows us to identify the initial few sentences during greedy decoding.


\item \textbf{RNN Decoder}. 
Another baseline is our proposed model without the encoder. 
The decoder hidden state is initialized with zeros.
We observed that using a special start symbol as for the other baselines helped obtain better performance with this model.
However, a start symbol did not help when the model is equipped with an encoder as the hidden state initialization alone was good enough.
\end{itemize}

We do not place emphasis on the particular search algorithm in this work and thus use beam search with the coherence score heuristic for all models.
A beam size of 100 was used. 
During decoding, sentence candidates that have been already chosen are pruned from the beam.
All RNNs use a hidden layer size of 1000.
For the window network we used a window size of 3 and a hidden layer size of 2000. 
We initialize all models with pre-trained GloVe word embeddings.

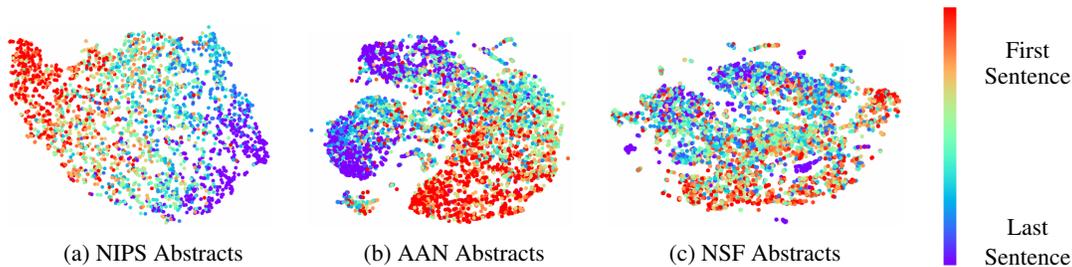
\begin{figure*}[t!]
	\centering
	\begin{subfigure}[b]{0.22\textwidth} 
		\includegraphics[width=0.9\textwidth]{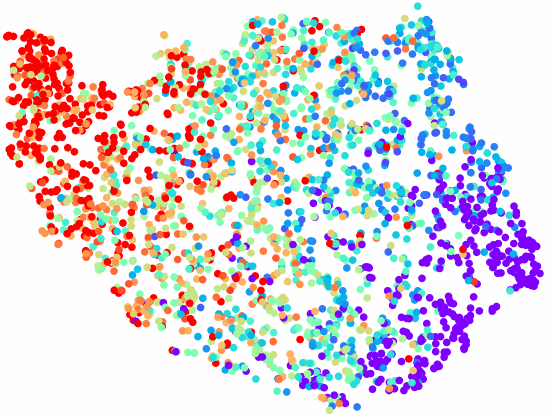}
		\caption{NIPS Abstracts}
	\end{subfigure}
	\begin{subfigure}[b]{0.22\textwidth} 
		\includegraphics[width=0.9\textwidth]{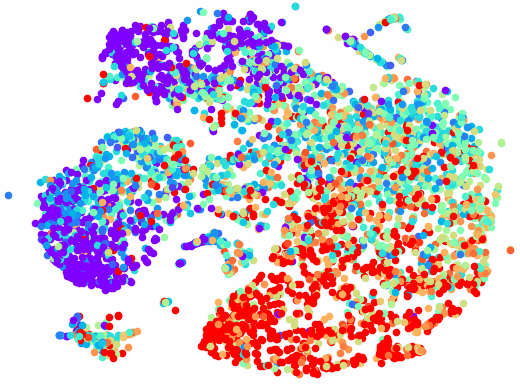}
		\caption{AAN Abstracts}
	\end{subfigure}
	\begin{subfigure}[b]{0.22\textwidth} 
		\includegraphics[width=\textwidth]{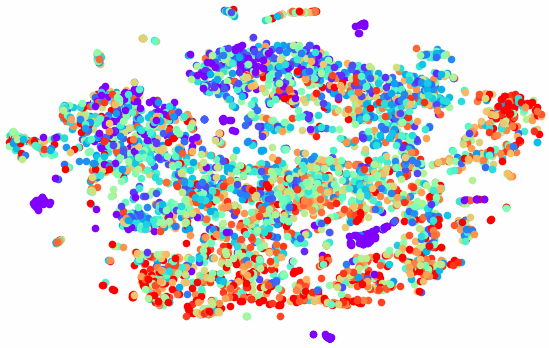}
		\caption{NSF Abstracts}
	\end{subfigure}
	\hspace{1em}
	\includegraphics[width=0.2\textwidth,angle=90]{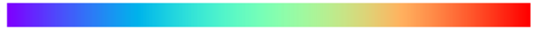} 
	\hspace{0.6em}
	\raisebox{0.4em}{
	\begin{tikzpicture}
		\node[anchor=north, inner sep=-3pt, outer xsep=-4pt]  (note) at (10,11.4) {\small First };
		\node[anchor=north, inner sep=-3pt, outer xsep=-4pt]  (note) at (10,11.05) {\small Sentence };
		\node[anchor=south, inner sep=-3pt, outer xsep=-4pt] (water) at (10,9) {\small Last };
		\node[anchor=north, inner sep=-3pt, outer xsep=-4pt]  (note) at (10,8.65) {\small Sentence };
	\end{tikzpicture}
	}
	\caption{t-SNE embeddings of representations learned by the model for sentences from the test set. Embeddings are color coded by the position of the sentence in the document it appears.}
	\label{tsne}
\end{figure*}

\begin{table*}[!t]
\parbox{.40\linewidth}{
\caption{ 
Comparison on extractive summarization between models trained from scratch and models pre-trained with the ordering task. 
}
\begin{center}
\begin{tabular}{l|c c c}
Model & {\scriptsize ROUGE-1} & {\scriptsize ROUGE-2} & {\scriptsize ROUGE-L} \\
\hline \\[-1.0em]
& \multicolumn{3}{c}{Summary length = 75b}  \\
From scratch            & 18.29 & 47.56 & 12.79 \\
Pre-train ord. & 18.77 & 50.32 & 13.25 \\
\hline \\[-1.0em]
& \multicolumn{3}{c}{Summary length = 275b}  \\
From scratch            & 35.82 & 10.67 & 33.69 \\
Pre-train ord.     & 36.47 & 10.99 & 34.27 \\

\end{tabular}
\end{center}
\label{rouge}
}
\hfill
\parbox{.58\linewidth}{
\caption{ 
Performance comparison for semantic similarity and paraphrase detection. 
The first row shows the best performing purely supervised methods. 
The last section shows our models.
}
\begin{center}
\begin{tabular}{l|c c c c c }
Model & \multicolumn{3}{c}{SICK} & \multicolumn{2}{c}{MSRP} \\ 
& r & $\rho$ & MSE & (Acc) & (F1) \\ 
\hline \\[-1.0em]
Supervised & 0.868 & 0.808 & 0.253 & 80.4 & 86.0 \\ 
Uni-ST	~\citep{kiros2015skip}	& 0.848 & 0.778 & 0.287 & 73.0 & 81.9 \\ 
\hline \\[-1.0em]
Ordering model	& 0.807 & 0.742 & 0.356 & 72.3 & 81.1 \\ 
+ BoW			& 0.842 & 0.775 & 0.299 & 74.0 & 81.9 \\ 
+ Uni-ST		& 0.860 & 0.795 & 0.270 & 74.9 & 82.5 \\ 
\end{tabular}
\end{center}
\label{sent_comp}
}
\end{table*}

We assess the performance of our model against baseline methods in Table \ref{nips}.
The window network performs strongly compared to the other baselines. 
Our model does better by a significant margin by exploiting global context, demonstrating that global context is important in this task. 

While the Entity-Grid model has been fairly successful for the order discrimination task in the past we observe that it fails to discriminate between a large number of candidates.
One reason could be that the feature representation is less sensitive to local changes in sentence order (such as swapping adjacent sentences).
The computational expense of obtaining parse trees and constructing grids on a large amount of data prohibited experimenting with this model on the NSF abstracts data.

The Seq2seq model performs worse than the window network.
Interestingly, \citet{li2016neural} observe that the Seq2seq model outperforms the window network in an order discrimination task on Wikipedia data.
However, the Wikipedia data considered in their work is an order of magnitude larger than the datasets considered here, and that could have potentially helped the generative model.
These models are also expensive during inference since they involve computing and sampling from word distributions.

Fig. \ref{tsne} shows t-SNE embeddings of sentence representations learned by our sentence encoder.
These are sentences from test sets, color coded by their positions in the source abstract.
This shows that our model learns high-level structure in the documents, generalizing well to unseen text.
The structure is less apparent in the NSF dataset due to its data diversity and longer documents.
While approaches based on the \citet{barzilay2004catching} model explicitly capture topics by discovering clusters in sentences, our neural approach implicitly discovers such structure.

\subsection{Sentence Ordering and Summarization}
In this section we show that sentence ordering models learn coherence properties useful for summarization.
We consider a variation of our model where the model takes a set of sentences from several documents as input and sequentially picks summary sentences until it predicts a special `stop' symbol. 
A key distinction between this model and recent work \citep{cheng2016neural,nallapati2016classify} is that the input order of sentences is assumed to be unknown, making it applicable to multi-document summarization. 

We train a model from scratch to perform extractive summarization in the above fashion.
We then consider a model that is pre-trained on the ordering task and is fine-tuned on the above task.
The DailyMail and CNN datasets \citep{cheng2016neural} were used for experimentation.
We use DailyMail for pre-training purposes and CNN for fine-tuning and evaluation.
The labels in DailyMail are not used.
We compare ROUGE scores of the two models in Table \ref{rouge} under standard evaluation settings.

We observe that the model pre-trained with the ordering task scores consistently better than the model trained from scratch.
The results can be improved further by using larger  news corpora. 
This shows that sentence ordering is an attractive unsupervised objective for exploiting large unlabelled corpora to improve summarization systems.
It further shows that the coherence scores obtained from the ordering model correlates well with summary quality.

\begin{table*}[!ht]
\centering
\caption{Visualizing salient words (Abstracts are from the AAN corpus). }
\begin{tabular}{p{0.97\textwidth}} 
\toprule
\textcolor[gray]{0.543}{In} \textcolor[gray]{0.605}{this} \textcolor[gray]{0.620}{paper} \textcolor[gray]{0.576}{,} \textcolor[gray]{0.413}{we} \textcolor[gray]{0.451}{propose} \textcolor[gray]{0.660}{a} \textcolor[gray]{0.453}{new} \textcolor[gray]{0.388}{method} \textcolor[gray]{0.445}{for} \textcolor[gray]{0.391}{semantic} \textcolor[gray]{0.409}{class} \textcolor[gray]{0.000}{induction} \textcolor[gray]{0.199}{.} \\
\textcolor[gray]{0.000}{First} \textcolor[gray]{0.379}{,} \textcolor[gray]{0.548}{we} \textcolor[gray]{0.398}{introduce} \textcolor[gray]{0.695}{a} \textcolor[gray]{0.540}{generative} \textcolor[gray]{0.714}{model} \textcolor[gray]{0.712}{of} \textcolor[gray]{0.637}{sentences} \textcolor[gray]{0.697}{,} \textcolor[gray]{0.693}{based} \textcolor[gray]{0.699}{on} \textcolor[gray]{0.583}{dependency} \textcolor[gray]{0.682}{trees} \textcolor[gray]{0.702}{and} \textcolor[gray]{0.682}{which} \textcolor[gray]{0.602}{takes} \textcolor[gray]{0.604}{into} \textcolor[gray]{0.423}{account} \textcolor[gray]{0.133}{homonymy} \textcolor[gray]{0.440}{.} \\
\textcolor[gray]{0.335}{Our} \textcolor[gray]{0.334}{model} \textcolor[gray]{0.290}{can} \textcolor[gray]{0.042}{thus} \textcolor[gray]{0.210}{be} \textcolor[gray]{0.000}{seen} \textcolor[gray]{0.464}{as} \textcolor[gray]{0.548}{a} \textcolor[gray]{0.199}{generalization} \textcolor[gray]{0.593}{of} \textcolor[gray]{0.317}{Brown} \textcolor[gray]{0.240}{clustering} \textcolor[gray]{0.479}{.} \\
\textcolor[gray]{0.000}{Second} \textcolor[gray]{0.444}{,} \textcolor[gray]{0.536}{we} \textcolor[gray]{0.479}{describe} \textcolor[gray]{0.659}{an} \textcolor[gray]{0.636}{efficient} \textcolor[gray]{0.643}{algorithm} \textcolor[gray]{0.674}{to} \textcolor[gray]{0.645}{perform} \textcolor[gray]{0.653}{inference} \textcolor[gray]{0.691}{and} \textcolor[gray]{0.621}{learning} \textcolor[gray]{0.602}{in} \textcolor[gray]{0.451}{this} \textcolor[gray]{0.358}{model} \textcolor[gray]{0.117}{.} \\
\textcolor[gray]{0.173}{Third} \textcolor[gray]{0.395}{,} \textcolor[gray]{0.467}{we} \textcolor[gray]{0.402}{apply} \textcolor[gray]{0.654}{our} \textcolor[gray]{0.620}{proposed} \textcolor[gray]{0.697}{method} \textcolor[gray]{0.720}{on} \textcolor[gray]{0.724}{two} \textcolor[gray]{0.708}{large} \textcolor[gray]{0.700}{datasets} \textcolor[gray]{0.755}{(} \textcolor[gray]{0.698}{108} \textcolor[gray]{0.737}{tokens} \textcolor[gray]{0.777}{,} \textcolor[gray]{0.752}{105} \textcolor[gray]{0.774}{words} \textcolor[gray]{0.768}{types} \textcolor[gray]{0.768}{)} \textcolor[gray]{0.754}{,} \textcolor[gray]{0.762}{and} \textcolor[gray]{0.741}{demonstrate} \textcolor[gray]{0.775}{that} \textcolor[gray]{0.747}{classes} \textcolor[gray]{0.710}{induced} \textcolor[gray]{0.748}{by} \textcolor[gray]{0.729}{our} \textcolor[gray]{0.677}{algorithm} \textcolor[gray]{0.698}{improve} \textcolor[gray]{0.728}{performance} \textcolor[gray]{0.716}{over} \textcolor[gray]{0.694}{Brown} \textcolor[gray]{0.658}{clustering} \textcolor[gray]{0.733}{on} \textcolor[gray]{0.737}{the} \textcolor[gray]{0.633}{task} \textcolor[gray]{0.665}{of} \textcolor[gray]{0.615}{semisupervised} \textcolor[gray]{0.592}{supersense} \textcolor[gray]{0.509}{tagging} \textcolor[gray]{0.640}{and} \textcolor[gray]{0.528}{named} \textcolor[gray]{0.466}{entity} \textcolor[gray]{0.241}{recognition} \textcolor[gray]{0.000}{.} \\
\cmidrule{1-1}
\textcolor[gray]{0.000}{Representation} \textcolor[gray]{0.415}{learning} \textcolor[gray]{0.305}{is} \textcolor[gray]{0.279}{a} \textcolor[gray]{0.177}{promising} \textcolor[gray]{0.538}{technique} \textcolor[gray]{0.598}{for} \textcolor[gray]{0.248}{discovering} \textcolor[gray]{0.428}{features} \textcolor[gray]{0.593}{that} \textcolor[gray]{0.639}{allow} \textcolor[gray]{0.626}{supervised} \textcolor[gray]{0.689}{classifiers} \textcolor[gray]{0.693}{to} \textcolor[gray]{0.569}{generalize} \textcolor[gray]{0.663}{from} \textcolor[gray]{0.741}{a} \textcolor[gray]{0.722}{source} \textcolor[gray]{0.735}{domain} \textcolor[gray]{0.689}{dataset} \textcolor[gray]{0.742}{to} \textcolor[gray]{0.680}{arbitrary} \textcolor[gray]{0.629}{new} \textcolor[gray]{0.559}{domains} \textcolor[gray]{0.626}{.} \\
\textcolor[gray]{0.095}{We} \textcolor[gray]{0.162}{present} \textcolor[gray]{0.511}{a} \textcolor[gray]{0.509}{novel} \textcolor[gray]{0.494}{,} \textcolor[gray]{0.058}{formal} \textcolor[gray]{0.526}{statement} \textcolor[gray]{0.508}{of} \textcolor[gray]{0.283}{the} \textcolor[gray]{0.120}{representation} \textcolor[gray]{0.245}{learning} \textcolor[gray]{0.014}{task} \textcolor[gray]{0.000}{.} \\
\textcolor[gray]{0.302}{We} \textcolor[gray]{0.000}{argue} \textcolor[gray]{0.435}{that} \textcolor[gray]{0.419}{because} \textcolor[gray]{0.633}{the} \textcolor[gray]{0.575}{task} \textcolor[gray]{0.592}{is} \textcolor[gray]{0.447}{computationally} \textcolor[gray]{0.511}{intractable} \textcolor[gray]{0.703}{in} \textcolor[gray]{0.551}{general} \textcolor[gray]{0.616}{,} \textcolor[gray]{0.520}{it} \textcolor[gray]{0.408}{is} \textcolor[gray]{0.302}{important} \textcolor[gray]{0.611}{for} \textcolor[gray]{0.635}{a} \textcolor[gray]{0.548}{representation} \textcolor[gray]{0.528}{learner} \textcolor[gray]{0.655}{to} \textcolor[gray]{0.651}{be} \textcolor[gray]{0.560}{able} \textcolor[gray]{0.663}{to} \textcolor[gray]{0.606}{incorporate} \textcolor[gray]{0.659}{expert} \textcolor[gray]{0.663}{knowledge} \textcolor[gray]{0.571}{during} \textcolor[gray]{0.615}{its} \textcolor[gray]{0.580}{search} \textcolor[gray]{0.555}{for} \textcolor[gray]{0.352}{helpful} \textcolor[gray]{0.367}{features} \textcolor[gray]{0.398}{.} \\
\textcolor[gray]{0.168}{Leveraging} \textcolor[gray]{0.493}{the} \textcolor[gray]{0.387}{Posterior} \textcolor[gray]{0.322}{Regularization} \textcolor[gray]{0.396}{framework} \textcolor[gray]{0.522}{,} \textcolor[gray]{0.425}{we} \textcolor[gray]{0.412}{develop} \textcolor[gray]{0.657}{an} \textcolor[gray]{0.623}{architecture} \textcolor[gray]{0.667}{for} \textcolor[gray]{0.444}{incorporating} \textcolor[gray]{0.491}{biases} \textcolor[gray]{0.594}{into} \textcolor[gray]{0.505}{representation} \textcolor[gray]{0.294}{learning} \textcolor[gray]{0.000}{.} \\
\textcolor[gray]{0.078}{We} \textcolor[gray]{0.000}{investigate} \textcolor[gray]{0.137}{three} \textcolor[gray]{0.548}{types} \textcolor[gray]{0.692}{of} \textcolor[gray]{0.437}{biases} \textcolor[gray]{0.639}{,} \textcolor[gray]{0.603}{and} \textcolor[gray]{0.549}{experiments} \textcolor[gray]{0.567}{on} \textcolor[gray]{0.545}{two} \textcolor[gray]{0.547}{domain} \textcolor[gray]{0.580}{adaptation} \textcolor[gray]{0.445}{tasks} \textcolor[gray]{0.572}{show} \textcolor[gray]{0.584}{that} \textcolor[gray]{0.521}{our} \textcolor[gray]{0.180}{biased} \textcolor[gray]{0.389}{learners} \textcolor[gray]{0.433}{identify} \textcolor[gray]{0.433}{significantly} \textcolor[gray]{0.557}{better} \textcolor[gray]{0.561}{sets} \textcolor[gray]{0.710}{of} \textcolor[gray]{0.537}{features} \textcolor[gray]{0.534}{than} \textcolor[gray]{0.419}{unbiased} \textcolor[gray]{0.423}{learners} \textcolor[gray]{0.669}{,} \textcolor[gray]{0.451}{resulting} \textcolor[gray]{0.652}{in} \textcolor[gray]{0.651}{a} \textcolor[gray]{0.438}{relative} \textcolor[gray]{0.599}{reduction} \textcolor[gray]{0.664}{in} \textcolor[gray]{0.574}{error} \textcolor[gray]{0.646}{of} \textcolor[gray]{0.548}{more} \textcolor[gray]{0.592}{than} \textcolor[gray]{0.493}{16}\textcolor[gray]{0.389}{\%} \textcolor[gray]{0.524}{for} \textcolor[gray]{0.496}{both} \textcolor[gray]{0.565}{tasks} \textcolor[gray]{0.648}{,} \textcolor[gray]{0.606}{with} \textcolor[gray]{0.525}{respect} 
\textcolor[gray]{0.626}{to} \textcolor[gray]{0.455}{state-of-the-art} \textcolor[gray]{0.479}{representation} \textcolor[gray]{0.427}{learning} \textcolor[gray]{0.109}{techniques}\textcolor[gray]{0.086}{.}\\
\bottomrule
\end{tabular}
\label{table:salient}
\end{table*}

\subsection{Learned Sentence Representations}
\label{senrep}

One of the original motivations for this work is the question of whether we can learn high-quality sentence representations by learning to model text coherence.
 To address this question we trained our model on a large number of paragraphs using the BookCorpus dataset~\citep{kiros2015skip}.

To evaluate the quality of sentence embeddings derived from the model, we use the evaluation pipeline of \citet{kiros2015skip} for tasks that involve understanding sentence semantics.
These evaluations are performed by training a classifier on top of the embeddings derived from the model (holding the embeddings fixed) so that the performance is indicative of the quality of sentence representations.
We present a comparison for the semantic relatedness and paraphrase detection tasks in Table \ref{sent_comp}.
Results for only uni-directional versions of models are discussed here for a fair comparison.
Similar to the skip-thought (ST) paper, we train two models - one predicting the correct order in the forward direction and another in the backward direction.
The numbers shown for the ordering model were obtained by concatenating the representations from the two models. 

Concatenating the above representation with the bag-of-words representation (using the fine-tuned word embeddings) of the sentence further improves performance. 
%
This is because the ordering model can choose to pay less attention to specific lexical information and focus on high-level document structure.
Hence, the two representations capture complementary semantics.
Adding ST features improves performance further.
We observed that the bilinear scoring function and introducing contrastive sentences to the decoder improved the quality of learned representations significantly. 

Our model has several key advantages over ST.
ST has a word-level reconstruction objective and is trained with large softmax output layers.
This limits the vocabulary size and slows down training (they use a vocabulary size of 20k and report two weeks of training).
Our model achieves comparable performance and does not have such a word reconstruction component.
We train with a vocabulary of 400k words; the above results are based on a training time of two days on a GTX Titan X GPU.

\subsection{Word Influence}
We attempt to understand what text-level clues the model uses to perform the ordering task. 
Inspired by \citet{li2015visualizing}, we use gradients of prediction decisions with respect to words of the correct sentence as a proxy for the salience of each word.
We feed sentences to the decoder in the correct order and at each time step compute the derivative of the score $e$ (Eq.~\ref{dec3}) of the correct next sentence $s = (w_1,..,w_n)$ with respect to its word embeddings. 
The importance of word $w_i$ in correctly predicting $s$ as the next sentence is defined as $\|\frac{\partial e}{\partial w_i}\|$.
We assume the hidden states of the decoder to be fixed and only back-propagate gradients through the sentence encoder.

Table \ref{table:salient} shows visualizations of two abstracts.
Darker shades correspond to higher gradient norms.
In the first example the model appears to be using the word clues `first', `second' and `third'. 
A similar observation was made by \citet{chen2016neural}. 
In the second example we observe that the model pays attention to phrases such as `We present', `We argue', which are typical of abstract texts.
It also focuses on the word `representation' appearing in the first two sentences.
These observations link to centering theory which states that entity distributions in coherent discourses exhibit certain patterns.
The model implicitly learns these patterns with no syntax annotations or handcrafted features.

\section{Conclusion}
This work investigated the challenging problem of coherently organizing a set of sentences.
Our RNN-based model performs strongly compared to baselines and prior work on sentence ordering and order discrimination tasks.
We further demonstrated that it captures high-level document structure and learns useful sentence representations when trained on large amounts of data.
Our approach to the ordering problem deviates from most prior work that use handcrafted features. 
However, exploiting linguistic features for next sentence classification can potentially further improve performance. 
Entity distribution patterns can provide useful features about named entities that are treated as out-of-vocabulary words.
The ordering problem can be further studied at higher-level discourse units such as paragraphs, sections and chapters.

\section{Acknowledgments}
This material is based in part upon work supported by IBM under contract 4915012629.
Any opinions, findings, conclusions or recommendations expressed above are those of the authors and do not necessarily reflect the views of IBM.
We thank the UMich/IBM Sapphire team and Junhyuk Oh, Ruben Villegas, Xinchen Yan, Rui Zhang, Kibok Lee and Yuting Zhang for helpful comments and discussions.

\bibliography{aaai}

\begin{thebibliography}{}

\bibitem[\protect\citeauthoryear{Barzilay and
  Elhadad}{2002}]{barzilay2002inferring}
Barzilay, R., and Elhadad, N.
\newblock 2002.
\newblock Inferring strategies for sentence ordering in multidocument news
  summarization.
\newblock {\em Journal of Artificial Intelligence Research}  35--55.

\bibitem[\protect\citeauthoryear{Barzilay and
  Lapata}{2008}]{barzilay2008modeling}
Barzilay, R., and Lapata, M.
\newblock 2008.
\newblock Modeling local coherence: An entity-based approach.
\newblock {\em Computational Linguistics} 34(1):1--34.

\bibitem[\protect\citeauthoryear{Barzilay and Lee}{2004}]{barzilay2004catching}
Barzilay, R., and Lee, L.
\newblock 2004.
\newblock Catching the drift: Probabilistic content models, with applications
  to generation and summarization.
\newblock {\em arXiv preprint cs/0405039}.

\bibitem[\protect\citeauthoryear{Burstein, Tetreault, and
  Andreyev}{2010}]{burstein2010using}
Burstein, J.; Tetreault, J.; and Andreyev, S.
\newblock 2010.
\newblock Using entity-based features to model coherence in student essays.
\newblock In {\em Human language technologies: The 2010 annual conference of
  the North American chapter of the Association for Computational Linguistics},
   681--684.
\newblock Association for Computational Linguistics.

\bibitem[\protect\citeauthoryear{Chen, Qiu, and Huang}{2016}]{chen2016neural}
Chen, X.; Qiu, X.; and Huang, X.
\newblock 2016.
\newblock Neural sentence ordering.
\newblock {\em arXiv preprint arXiv:1607.06952}.

\bibitem[\protect\citeauthoryear{Cheng and Lapata}{2016}]{cheng2016neural}
Cheng, J., and Lapata, M.
\newblock 2016.
\newblock Neural summarization by extracting sentences and words.
\newblock {\em arXiv preprint arXiv:1603.07252}.

\bibitem[\protect\citeauthoryear{Doersch, Gupta, and
  Efros}{2015}]{doersch2015unsupervised}
Doersch, C.; Gupta, A.; and Efros, A.~A.
\newblock 2015.
\newblock Unsupervised visual representation learning by context prediction.
\newblock In {\em Proceedings of the IEEE International Conference on Computer
  Vision},  1422--1430.

\bibitem[\protect\citeauthoryear{Elsner, Austerweil, and
  Charniak}{2007}]{elsner2007unified}
Elsner, M.; Austerweil, J.~L.; and Charniak, E.
\newblock 2007.
\newblock A unified local and global model for discourse coherence.
\newblock In {\em HLT-NAACL},  436--443.

\bibitem[\protect\citeauthoryear{Grosz, Weinstein, and
  Joshi}{1995}]{grosz1995centering}
Grosz, B.~J.; Weinstein, S.; and Joshi, A.~K.
\newblock 1995.
\newblock Centering: A framework for modeling the local coherence of discourse.
\newblock {\em Computational linguistics} 21(2):203--225.

\bibitem[\protect\citeauthoryear{Guinaudeau and
  Strube}{2013}]{guinaudeau2013graph}
Guinaudeau, C., and Strube, M.
\newblock 2013.
\newblock Graph-based local coherence modeling.
\newblock In {\em ACL (1)},  93--103.

\bibitem[\protect\citeauthoryear{Kiddon, Zettlemoyer, and
  Choi}{2016}]{kiddon2016globally}
Kiddon, C.; Zettlemoyer, L.; and Choi, Y.
\newblock 2016.
\newblock Globally coherent text generation with neural checklist models.
\newblock In {\em Proceedings of the 2016 Conference on Empirical Methods in
  Natural Language Processing (EMNLP)}.

\bibitem[\protect\citeauthoryear{Kingma and Ba}{2014}]{kingma2014adam}
Kingma, D., and Ba, J.
\newblock 2014.
\newblock Adam: A method for stochastic optimization.
\newblock {\em arXiv preprint arXiv:1412.6980}.

\bibitem[\protect\citeauthoryear{Kiros \bgroup et al\mbox.\egroup
  }{2015}]{kiros2015skip}
Kiros, R.; Zhu, Y.; Salakhutdinov, R.~R.; Zemel, R.; Urtasun, R.; Torralba, A.;
  and Fidler, S.
\newblock 2015.
\newblock Skip-thought vectors.
\newblock In {\em Advances in Neural Information Processing Systems},
  3276--3284.

\bibitem[\protect\citeauthoryear{Klein and Manning}{2003}]{klein2003accurate}
Klein, D., and Manning, C.~D.
\newblock 2003.
\newblock Accurate unlexicalized parsing.
\newblock In {\em Proceedings of the 41st Annual Meeting on Association for
  Computational Linguistics-Volume 1},  423--430.
\newblock Association for Computational Linguistics.

\bibitem[\protect\citeauthoryear{Lapata}{2003}]{lapata2003probabilistic}
Lapata, M.
\newblock 2003.
\newblock Probabilistic text structuring: Experiments with sentence ordering.
\newblock In {\em Proceedings of the 41st Annual Meeting on Association for
  Computational Linguistics-Volume 1},  545--552.
\newblock Association for Computational Linguistics.

\bibitem[\protect\citeauthoryear{Lapata}{2006}]{lapata2006automatic}
Lapata, M.
\newblock 2006.
\newblock Automatic evaluation of information ordering: Kendall's tau.
\newblock {\em Computational Linguistics} 32(4):471--484.

\bibitem[\protect\citeauthoryear{Li and Hovy}{2014}]{li2014model}
Li, J., and Hovy, E.~H.
\newblock 2014.
\newblock A model of coherence based on distributed sentence representation.
\newblock In {\em EMNLP},  2039--2048.

\bibitem[\protect\citeauthoryear{Li and Jurafsky}{2016}]{li2016neural}
Li, J., and Jurafsky, D.
\newblock 2016.
\newblock Neural net models for open-domain discourse coherence.
\newblock {\em arXiv preprint arXiv:1606.01545}.

\bibitem[\protect\citeauthoryear{Li \bgroup et al\mbox.\egroup
  }{2015}]{li2015visualizing}
Li, J.; Chen, X.; Hovy, E.; and Jurafsky, D.
\newblock 2015.
\newblock Visualizing and understanding neural models in nlp.
\newblock {\em arXiv preprint arXiv:1506.01066}.

\bibitem[\protect\citeauthoryear{Li, Luong, and
  Jurafsky}{2015}]{li2015hierarchical}
Li, J.; Luong, M.-T.; and Jurafsky, D.
\newblock 2015.
\newblock A hierarchical neural autoencoder for paragraphs and documents.
\newblock {\em arXiv preprint arXiv:1506.01057}.

\bibitem[\protect\citeauthoryear{Lichman}{2013}]{Lichman:2013}
Lichman, M.
\newblock 2013.
\newblock {UCI} machine learning repository.

\bibitem[\protect\citeauthoryear{Lin \bgroup et al\mbox.\egroup
  }{2015}]{lin2015hierarchical}
Lin, R.; Liu, S.; Yang, M.; Li, M.; Zhou, M.; and Li, S.
\newblock 2015.
\newblock Hierarchical recurrent neural network for document modeling.
\newblock In {\em Proceedings of the 2015 Conference on Empirical Methods in
  Natural Language Processing},  899--907.

\bibitem[\protect\citeauthoryear{Louis and Nenkova}{2012}]{louis2012coherence}
Louis, A., and Nenkova, A.
\newblock 2012.
\newblock A coherence model based on syntactic patterns.
\newblock In {\em Proceedings of the 2012 Joint Conference on Empirical Methods
  in Natural Language Processing and Computational Natural Language Learning},
  1157--1168.
\newblock Association for Computational Linguistics.

\bibitem[\protect\citeauthoryear{Miltsakaki and
  Kukich}{2004}]{miltsakaki2004evaluation}
Miltsakaki, E., and Kukich, K.
\newblock 2004.
\newblock Evaluation of text coherence for electronic essay scoring systems.
\newblock {\em Natural Language Engineering} 10(01):25--55.

\bibitem[\protect\citeauthoryear{Nallapati, Zhou, and
  Ma}{2016}]{nallapati2016classify}
Nallapati, R.; Zhou, B.; and Ma, M.
\newblock 2016.
\newblock Classify or select: Neural architectures for extractive document
  summarization.
\newblock {\em arXiv preprint arXiv:1611.04244}.

\bibitem[\protect\citeauthoryear{Nguyen and Joty}{2017}]{nguyen2017neural}
Nguyen, D.~T., and Joty, S.
\newblock 2017.
\newblock A neural local coherence model.
\newblock In {\em Proceedings of the 55th Annual Meeting of the Association for
  Computational Linguistics (Volume 1: Long Papers)}, volume~1,  1320--1330.

\bibitem[\protect\citeauthoryear{Noroozi and
  Favaro}{2016}]{noroozi2016unsupervised}
Noroozi, M., and Favaro, P.
\newblock 2016.
\newblock Unsupervised learning of visual representations by solving jigsaw
  puzzles.
\newblock In {\em European Conference on Computer Vision},  69--84.
\newblock Springer.

\bibitem[\protect\citeauthoryear{Park and Kim}{2015}]{park2015expressing}
Park, C.~C., and Kim, G.
\newblock 2015.
\newblock Expressing an image stream with a sequence of natural sentences.
\newblock In {\em Advances in Neural Information Processing Systems},  73--81.

\bibitem[\protect\citeauthoryear{Pennington, Socher, and
  Manning}{2014}]{pennington2014glove}
Pennington, J.; Socher, R.; and Manning, C.~D.
\newblock 2014.
\newblock Glove: Global vectors for word representation.
\newblock In {\em EMNLP}, volume~14,  1532--43.

\bibitem[\protect\citeauthoryear{Radev \bgroup et al\mbox.\egroup
  }{2009}]{Radev&al.09}
Radev, D.~R.; Joseph, M.~T.; Gibson, B.; and Muthukrishnan, P.
\newblock 2009.
\newblock {A} {B}ibliometric and {N}etwork {A}nalysis of the field of
  {C}omputational {L}inguistics.
\newblock {\em Journal of the American Society for Information Science and
  Technology}.

\bibitem[\protect\citeauthoryear{Soricut and
  Marcu}{2006}]{soricut2006discourse}
Soricut, R., and Marcu, D.
\newblock 2006.
\newblock Discourse generation using utility-trained coherence models.
\newblock In {\em Proceedings of the COLING/ACL on Main conference poster
  sessions},  803--810.
\newblock Association for Computational Linguistics.

\bibitem[\protect\citeauthoryear{Sutskever, Vinyals, and
  Le}{2014}]{sutskever2014sequence}
Sutskever, I.; Vinyals, O.; and Le, Q.~V.
\newblock 2014.
\newblock Sequence to sequence learning with neural networks.
\newblock In {\em Advances in neural information processing systems},
  3104--3112.

\bibitem[\protect\citeauthoryear{Vinyals, Bengio, and
  Kudlur}{2015}]{vinyals2015order}
Vinyals, O.; Bengio, S.; and Kudlur, M.
\newblock 2015.
\newblock Order matters: Sequence to sequence for sets.
\newblock {\em arXiv preprint arXiv:1511.06391}.

\bibitem[\protect\citeauthoryear{Vinyals, Fortunato, and
  Jaitly}{2015}]{vinyals2015pointer}
Vinyals, O.; Fortunato, M.; and Jaitly, N.
\newblock 2015.
\newblock Pointer networks.
\newblock In {\em Advances in Neural Information Processing Systems},
  2674--2682.

\bibitem[\protect\citeauthoryear{Wang and Gupta}{2015}]{wang2015unsupervised}
Wang, X., and Gupta, A.
\newblock 2015.
\newblock Unsupervised learning of visual representations using videos.
\newblock In {\em Proceedings of the IEEE International Conference on Computer
  Vision},  2794--2802.

\end{thebibliography}
\bibliographystyle{aaai}
\end{document}